\documentclass[10pt,twocolumn,letterpaper]{article}
\usepackage{cvpr}

\usepackage{graphicx}
\usepackage{amsmath}
\usepackage{amssymb}
\usepackage{booktabs}
\usepackage{algorithm}
\usepackage{algpseudocode}
\floatname{algorithm}{Procedure}

\usepackage[pagebackref,breaklinks,colorlinks]{hyperref}

\usepackage[capitalize]{cleveref}
\crefname{section}{Sec.}{Secs.}
\Crefname{section}{Section}{Sections}
\Crefname{table}{Table}{Tables}
\crefname{table}{Tab.}{Tabs.}
\usepackage{pifont}
\newcommand{\cmark}{\ding{51}}%
\newcommand{\xmark}{\ding{55}}%

\def\cvprPaperID{6392}
\def\confName{CVPR}
\def\confYear{2023}

\begin{document}

\title{Chart-RCNN: Efficient Line Chart Data Extraction from Camera Images}

\author{
    Shufan Li\\
	Orka Labs Inc.\\
	\texttt{lishufan@hiorka.com}
	\and
	Congxi Lu\\
	Orka Labs Inc.\\
	\texttt{chauncey@hiorka.com} \\
	\and
	Linkai Li\\
	Orka Labs Inc.\\
	\texttt{linkai@hiorka.com} \\
	\and
	Haoshuai Zhou\thanks{Corresponding Author}\\
	Orka Labs Inc.\\
	\texttt{haoshuai@hiorka.com}
}
\maketitle

\begin{abstract}
Line Chart Data Extraction is a natural extension of Optical Character Recognition where the objective is to recover the underlying numerical information a chart image represents. Some recent works such as ChartOCR approach this problem using multi-stage networks combining OCR models with object detection frameworks. However, most of the existing datasets and models are based on "clean" images such as screenshots that drastically differ from camera photos. In addition, creating domain-specific new datasets requires extensive labeling which can be time-consuming. Our main contributions are as follows: we propose a synthetic data generation framework and a one-stage model that outputs text labels, mark coordinates, and perspective estimation simultaneously. We collected two datasets consisting of real camera photos for evaluation. Results show that our model trained only on synthetic data can be applied to real photos without any fine-tuning and is feasible for real-world application.
\end{abstract}
\begin{figure}[!t]
    \centering
    \includegraphics[width=0.8\linewidth]{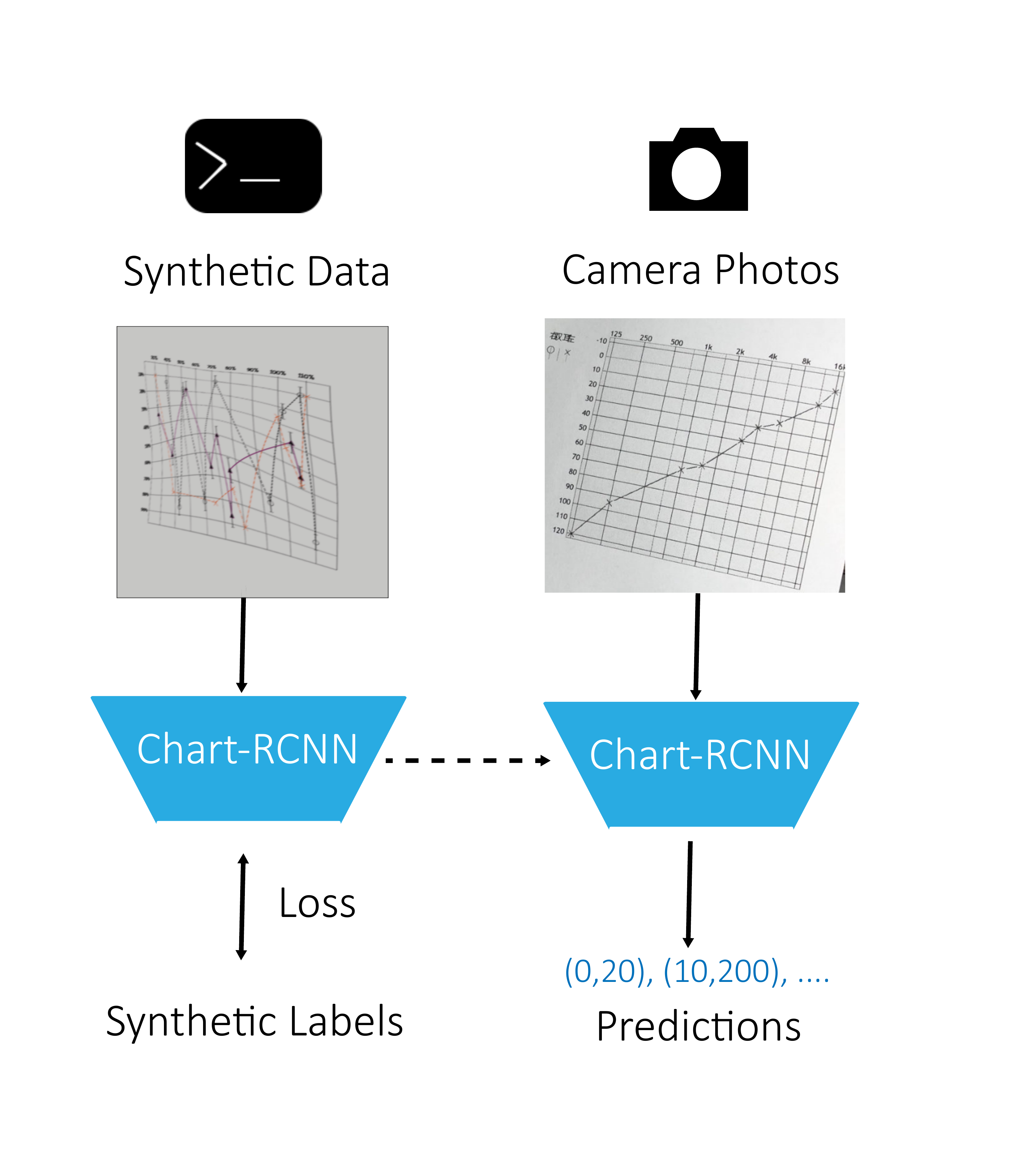}
    \caption{\textbf{An overview of our method. } We train our model on synthetic data of chart images and perform inference on real camera photos. Our main contributions are 1. A synthetic dataset pipeline. 2. A benchmark dataset of camera photos. 3. An end-to-end trainable model. }
    \label{fig:teaser}
\end{figure}
\section{Introduction}
\label{sec:intro}
Line charts are widely used to represent data in a wide range of documents such as financial forecasts, research publications, and medical reports \cite{kobayashi1999toward,wang2003cost}. Automatic analysis of line charts can facilitate document mining,  medical diagnosis, and help the visually impaired understand documents. However, when line charts are published in images, the raw data is lost. Recovering the underlying information of line charts will improve the performance of existing chart classification and question-answering systems such as \cite{kafle2018dvqa,Kahou2018FigureQAAA,kosemen2020multi}. It is trivial to identify the maximum value in a line chart given the raw data, but not so given only an image. Additionally, some analysis can only be performed when all the raw data is available. For example, when an audiologist fits a hearing aid for a patient given an audiogram ( a line chart representing one's hearing loss level at different frequencies), he or she must read the exact values from the chart. Therefore, Line Chart Data Extraction is a task of great significance as it improves the accuracy of qualitative analysis such as classification and question-answering, and makes quantitative applications such as automatic hearing-aid fitting possible. 

There have been numerous methods \cite{huang2007extraction,zhou2001chart} based on manually designed features and rules, however, they failed to generalize a diverse range of line chart designs. Many recent works such as \cite{Kato2022ParsingLC,luo2021chartocr} achieved better performance using deep neural networks. But these systems are difficult to train in that the input image is typically passed through a sequence of independent networks performing different tasks such as OCR and object detection, each requiring a separate training pipeline. 

One major limitation of these works is that they focus on extracting data from "clean" chart images such as scrapped images from the Internet, Excel sheets, and PDF documents \cite{Kato2022ParsingLC,luo2021chartocr,savva2011revision}. We argue that extracting chart data from real-world camera photos is equally important, if not more. There is an analogy between scene text recognition as an extension of optical character recognition(OCR) and chart data extraction from camera images as an extension of chart data extraction from "clean" images. Just as scene text recognition powers real-time documentation translation and text-to-speech services, chart data extraction from camera images can power applications such as helping the visually impaired to read charts in printed documents and automatic tuning of hearing aids using audiograms photos taken by cellphone cameras. Unfortunately, this natural extension of chart data extraction hasn't been fully explored.


We believe there are several reasons that this area is currently under-explored. First, Deep Neural Networks can heavily overfit small datasets, which is why current deep methods such as \cite{luo2021chartocr,Kato2022ParsingLC} are trained on large-scale datasets of scrapped or generated images. However, there are no large-scale datasets of camera photos of line charts available. Additionally, collecting and annotating such photos can be costly and time-consuming. Second, existing methods rely on pre-trained OCR models.  Camera photos introduce additional skew, rotation, and color perturbations to texts which lead to more noise in OCR results and make the task more challenging. Third, existing methods rely on bounding box estimations to find chart regions and assume the x and y-axis of the chart are perfectly aligned with that of the image. However, in real camera photos, the chart is projected onto the image plane through a homography transformation, which means additional components need to be added to the system to predict such transformation. Given many of these systems are already very complicated, this is no trivial task. 

In this paper, we tackle the problem of chart data extraction from camera images by directly addressing these challenges. Our strategy is to develop a system that pretrains on synthetic data only but can perform inference on real camera photos. \cref{fig:teaser} is an illustration of our proposed system. Our main contribution can be summarized as follows. 1) We propose Chart-RCNN, an end-to-end trainable network that extracts line chart data from camera photos. 2) We propose a synthetic data generation pipeline that can be used to train models capable of performing inference in real camera photos. 3) We collected datasets of camera photos of both real and synthetic charts with annotations of raw data. These datasets can be used to benchmark the performance of future works in this area

\section{Related Works}
\label{sec:related_works}

\subsection{Line Chart Datasets }
\textbf{PMC} \cite{Davila2020ICPR2} dataset contains real charts extracted from Open-Access publications found in the PubMedCentral. (PMC). It contains 7,401 and 3,155 line charts in the training and test sets respectively. However, only 1486 images contain annotations of raw data. 

\textbf{FigureQA} \cite{Kahou2018FigureQAAA} is a synthetic dataset of over 100,000 images originally purposed for question-answering tasks. They also include the metadata used during the generation which makes it possible to use it as a chart data extraction dataset. However, they used only five line styles and the charts do not have any marks on them.

\textbf{ExcelChart400K} is published alongside ChartOCR \cite{luo2021chartocr}. It contains Images generated via crawled Excel sheets and annotation of keypoints. However, they do not provide ground truth OCR annotation nor the raw data used to generate the line charts. 

All these datasets contain only clean images. Our datasets on the contrary contain real camera photos. We also provide annotations of raw data and a diverse range of chart styles.

\subsection{Line Chart Data Extraction}
In typical rule-based systems such as \cite{10.1145/1284420.1284427,Poco2017ReverseEngineeringVR}, histogram-based techniques and spatial filtering are used to locate coordinate axes and data points, while OCR is employed to detect textblocks. Then a set of handicraft algorithms is applied to filter false detections and generate a final output. However, these algorithms fail to generalize to the diverse range of chart designs. To address this issue, some systems such as ChartSense \cite{Jung2017ChartSenseID} introduce an interactive process where a human agent can make corrections when necessary. However, this makes the data extraction process time-consuming and difficult to scale.

Deep-learning-based methods such as \cite{luo2021chartocr,Kato2022ParsingLC} outperform rule-based methods by training on large-scale datasets. ChartOCR \cite{luo2021chartocr} uses CornerNet to find keypoint proposals and Microsoft OCR API to perform text recognition. They also include a separate QUERY network to predict if two points belong to the same line. \cite{Kato2022ParsingLC} uses a segmentation network that produces segmentation masks for different lifestyles and uses linear programming to perform line tracing. Text recognition is performed using STAR-Net. However, these methods assume clean, aligned images as input and cannot handle camera photos. Additionally, unlike our methods which train an OCR network alongside the detection network, they rely on generic OCR models that do not take into account the prior distributions of tick numbers.

\section{Dataset}
\begin{table*}[t]
  \centering
  \begin{tabular}{ccccccc}
    \toprule
    Dataset & Image Source &  Raw Data & Style & Alignment  & Background\\
    \midrule
    PMC \cite{Davila2020ICPR2} &  Crawled &  Partial*  & Many  & Aligned & White \\
    FigureQA \cite{Kahou2018FigureQAAA}& Generated &     \cmark   & 5 styles   & Aligned  & White    \\
    ExcelChart400K\cite{luo2021chartocr} & Crawled &      \xmark & Many  & Aligned  & White  \\\hline
    \textbf{SYN-Train }& Generated &  \cmark  & 30+ styles  & Perspective & Synthetic \\
    \textbf{SYN-Camera} & Camera Photo &  \cmark  & 30+  styles& Perspective & Print \\
    \textbf{Audiogram} & Camera Photo &  \cmark  & 2  styles & Perspective & Print \\
    \textbf{Audiogram (Scanned)} & Scanned &  \cmark  & 2 styles & Aligned  & Print \
    \\\bottomrule
  \end{tabular}
  \caption{A comparison between our dataset and existing ones. The most significant difference is that our dataset contains more than clean images that are crawled or generated.}
  \label{tab:dataset_comparison}
\end{table*}
\begin{figure}[!t]
    \centering
    \includegraphics[width=\linewidth]{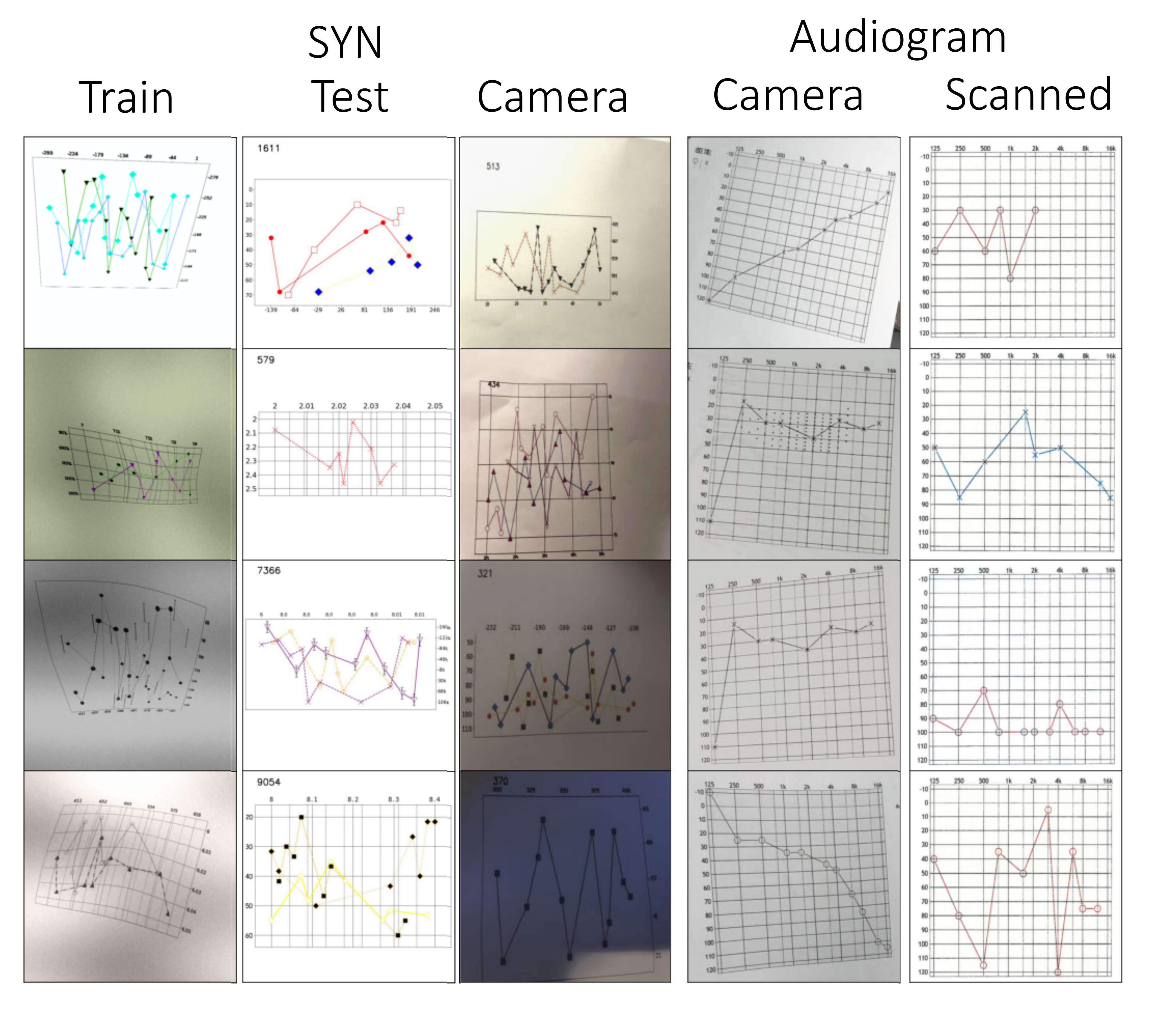}
    \caption{\textbf{Samples from Our Proposed Dataset.} SYN-Train is an online synthetic dataset with aggressive augmentation to mimic the look of camera photos. SYN-Test is a dataset of pretrained images for evaluation. SYN-Camera are real camera photos of charts in SYN-Test. Audiogram datasets consist of data generated by Noah 4, a software widely used by audiologists. There is a camera version and a scanned version. }
    \label{fig:dataset}
\end{figure}
We propose two datasets with variants. \textbf{SYN} dataset consists of synthetic data generated using matplotlib \cite{hunter2007matplotlib}. It has various mark and line styles. \textbf{SYN-Train} is an online dataset with infinitely many images. \textbf{SYN-Test} is a fixed subset of \textbf{SYN-Train} with 4,000 images. \textbf{SYN-Camera} contains 200 camera photos of charts in \textbf{SYN-Test}. \textbf{Audiogram} dataset consists of 420 camera photos of audiogram created using Noah 4\cite{himsa_noah}, a software system widely used in the hearing care industry for such purposes. 223 images have ground truth annotation of bounding boxes of labels and marks. This dataset is used to evaluate how our model performs in real-world scenarios and compare models trained on synthetic datasets with those trained on manually labeled data. Samples from all these datasets are shown in \cref{fig:dataset}.

Our datasets differ from existing ones mentioned in \cref{sec:related_works} in that they consist of camera photos or images that are visually similar to camera photos with annotations of raw data. \textbf{SYN} also have a larger variety in line styles compared with existing synthetic datasets such as \cite{Kahou2018FigureQAAA}. \cref{tab:dataset_comparison} illustrates a comparison between our datasets and existing ones.
\begin{figure*}
    \centering
    \includegraphics[width=1.0\textwidth]{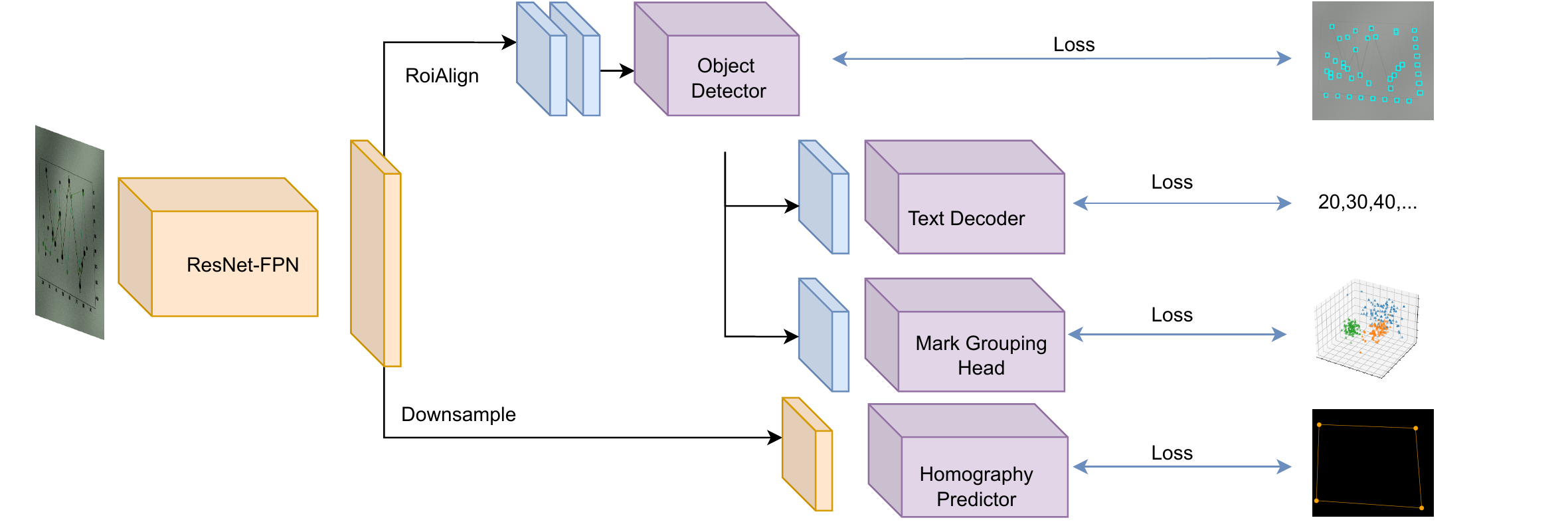}

   \caption{\textbf{Proposed Chart-RCNN architecture.} A standard Faster-RCNN with ResNet-FPN backbone is used to detect tick labels and marks, the out put is then passed to Text Decoder and Mark Grouping head for text recognition and mark clustering. A downsampled feature map from FPN is passed to Homography Prediction head to predict the perspective transforms of the image.}
   \label{fig:architecture}
\end{figure*}

\subsection{Synthetic }
\label{sec:synthetic_data_generation}
We use matplotlib to randomly generate line charts for \textbf{SYN} dataset. The images have a dimension of 720x720. Each of the axes has a random number of ticks between 5 and 10. The tick labels are randomly selected from a set of common intervals such as [0, 1], [0, 100], and [1000, 10000]. The stride between two consecutive labels is randomly selected among a set of common values such as 0.1, 0.2, 1, 10, and 100. Numbers less than 1 are randomly converted to percentile. Numbers larger than 1000 are randomly converted to human-readable notations such as 5k, and 10k. The grid lines are randomly added.  Each line chart has a random number of lines between 1 and 3 with random combinations of mark types, mark colors, and line styles. \cref{tab:syn_generation} illustrates all the parameters used in data generation.

After a clean image is generated, we apply a set of aggressive augmentations to make it visually similar to a camera photo. We first apply random Thin-Plate-Spline (TPS) \cite{tps} using a random grid to create small-scale permutations. We then apply a random perspective transform with a distortion scale of 0.5. We add random Gaussian noise at multiple scales.  We randomly apply color jittering to the resulting image, and finally, randomly add a blur effect using Gaussian Blur or Motion Blur. Figure \cref{fig:transforms} is an illustration of this pipeline. As shown in the figure, the clean image looks visually similar to a camera photo after these transformations. 

We generate 4,000 clean images without transforms as \textbf{SYN-Test} dataset. We print 200 of them on A4 paper and then took camera photos as the \textbf{SYN-Camera} dataset.

We propose two datasets with variants. \textbf{SYN} dataset consists of synthetic data generated using matplotlib . It has various mark and line styles. \textbf{SYN-Train} is an online dataset with infinitely many images. \textbf{SYN-Test} is a fixed subset of \textbf{SYN-Train} with 4,000 images. \textbf{SYN-Camera} contains 200 camera photos of charts in \textbf{SYN-Test}. \textbf{Audiogram} dataset consists of 420 camera photos of audiogram created using Noah 4 \cite{himsa_noah}, a software system widely used in the hearing care industry for such purposes. 223 images have ground truth annotation of bounding boxes of labels and marks. This dataset is used to evaluate how our model performs in real-world scenarios and compare models trained on synthetic datasets with those trained on manually labeled data. 

All these datasets have raw data available. All except for \textbf{SYN-Camera} also have annotations of bounding boxes of labels and marks, alongside their values.
a
\begin{table}
  \centering
  \begin{tabular}{cc}
    \toprule
    Parameter & Range\\
    \midrule
    Number of Ticks &  5-10 \\
    Number of Lines &  1-3 \\
    Grid &  on, off  \\
    Aspect Ratio & 0.25-1.0\\
    Mark Style & circle, triangle, cross, diamond, ... \\
    Line Style & solid, dot, dash, dash-dot\\
    Line Color & random \\
    \bottomrule
  \end{tabular}
  \caption{Randomized Parameters used in data generation. }
  \label{tab:syn_generation}
\end{table}

\subsection{Audiogram}

We first collected raw audiogram images generated by Noah 4 System used by audiologists. These audiograms were printed on standard A4-sized paper. We took a total of 420
images of these printed audiograms under various angles and lighting conditions as the \textbf{Audiogram} dataset. In some photos, there are cast shadows of other objects on the paper. In some photos, obstructions such as pens are purposely placed on the audiogram region. In some photos, the paper is bent or folded. These intentional occlusions and distortions make the dataset particularly
challenging. To evaluate the impact of these distortions, we generated an additional set of 30 audiograms, printed them,
and scanned them using a laser scanner to form \textbf{Audiogram-Scanned} Dataset.

All audiograms have their raw data available. We picked 223 camera photos and manually added annotations of ground truth bounding boxes of labels and marks.

\section{Method}
\label{fig:method}

We propose Chart-RCNN, an extension of Faster-RCNN \cite{Ren2015FasterRT} with additional heads attached to the ResNet-FPN backbone performing OCR, mark grouping and homography prediction. The network is trained jointly end-to-end. As is shown in \cref{fig:architecture}, an input image first go through standard Faster-RCNN blocks which generate bounding box predictions and classification for labels and marks. We use RoI Align to extract features in regions corresponding to text labels and marks, which are then passed to Text Decoder for OCR and Mark Grouping Head for clustering. The whole feature map from FPN backbone is downsampled and passed to a Homography Predictor to predict the perspective transforms. 
\begin{figure}[!t]
    \centering
    \includegraphics[width=\linewidth]{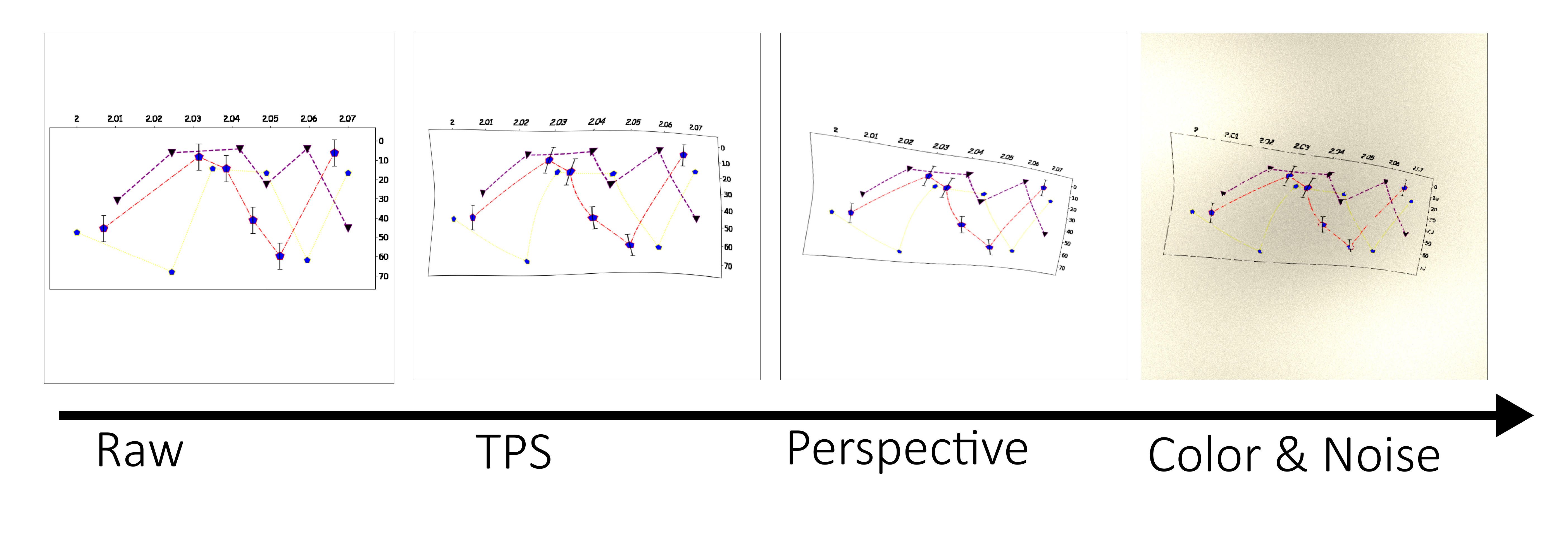}
    \caption{\textbf{Transforms of SYN Dataset}. We visualize transformed used in data generation process. TPS and Perspective produce local and global geometric distortions respectively, while the Color and Noise change the visual appearance of the image. }
    \label{fig:transforms}
\end{figure}
\subsection{Chart Element Detection}
We employ the standard Faster-RCNN head following \cite{Ren2015FasterRT}. The RoIAlign generates a 7x7 feature map which is passed to a Two-Layer MLP with a hidden dimension of 1024. The output is then passed to a linear layer for classification. We employ cross-entropy loss on the classification output.
\subsection{Homography Estimation}
\label{sec:homography}
We estimate the perspective transformation of an image by detecting four corners of the chart region. These four corners uniquely determine a projective transform that sends these four points respective to (0,0), (0,1), (1,1), (1,0) in the canonical coordinate system. The parameter of such a transformation can be determined by solving a linear system following \cite{mou2013robust} We explore two approaches to predicting the corners: Mask-based ones and keypoint-based ones. 

The mask-based approach pass the downsample feature maps to an FCN head \cite{Shelhamer2015FullyCN} consisting of two Convolution layers of size 3 and stride 1 with RELU activation, followed by a 1x1 convolution for pixel-wise classification.  We extract polygon contours from the mask image following \cite{Suzuki1985TopologicalSA}. The polygon is subsequently reduced to a quadrilateral using a modified version of Douglas‐Peucker Algorithm \cite{Visvalingam1990TheDA} (See appendix for more implementation details). 

The keypoint-based approach directly predicts the coordinates of four points following the design of Keypoint-RCNN \cite{He2020MaskR}. The architecture is similar to the mask-based approach. The only difference is that the prediction target is a 4-channel heatmap corresponding to 4 corners. 

We employ cross-entropy loss in both options.

\subsection{Text Recognition}
\label{sec:text_recog}
We explore two architecture for OCR: Transformer and CRNN. For Transformer, we followed the design of TrOCR\cite{Li2021TrOCRTO}. RoiAlign generates a feature map of 3x9 for each text region, which are then treated as 3x9 patches and passed to the Transformer block. For CRNN, we follow the design of \cite{Shi2017AnET}. The text regions are pooled to a 32x96 feature map. The corresponding RGB information of the original image is added to this feature map through a Residual connection. We employ cross entropy loss for Transformer and connectionist temporal classification(CTC) loss for CRNN.

\subsection{Mark Clustering}
The Mark Grouping Head projects the features to a space where the embedding of the marks from the same line are close to each other. It consists of a series of convolution blocks followed by a 2-layer MLP (See appendix for more implementation details). We then calculate the cosine similarity of all mark features. We assume the likelihood of two points belonging to the same line is defined by
\begin{equation}
\mathbb{P}(G(x,y)=1)=exp(\frac{f(x)^Tf(y)}{\tau\lVert f(x) \rVert \lVert f(y) \rVert })
\label{eq:likelyhood}
\end{equation}

Where $G(x,y)$ is an indicator function that equals to 1 if and only if mark $x$ and $y$ belongs to the same line. $f$ is the mark grouping head. $\tau$ is a hyperparameter. 

This formulation transforms the clustering problem to a binary classification problem and naturally leads to the  following loss:

\begin{multline}
\mathcal{L}_{Group}= - \sum_{i \ne j} [G(x_i,x_j) log \mathbb{P}(G(x,y)=1) \\ 
+ log(1- \mathbb{P}(G(x_i,x_j)=1))]
\label{eq:loss}
\end{multline}

The final loss of the entire Chart-RCNN model is simply the sum of loss of each module. 

\subsection{Post Processing}

We gather the output from different modules. We first remap the detected label and mark coordinates to the canonical coordinate system with four corners of the chart remapped to (0,0), (0,1), (1,1), and (1,0). We group the labels by their proximity to chart borders. Labels close to the top and bottom border are considered as labels for the x-axis, whereas those close to the left and right border are considered as labels for the y-axis. We retrieve the text output from Text Decoder and convert them to floats. Then we fit two RANSAC linear models to obtain a transformation from the canonical coordinate system to the raw data values. We apply such transformations to the coordinates of detected maps to get the raw data points. When necessary, we perform threshold-based hierarchical clustering \cite{rai2010survey} on mark features generated from the Mask head. When evaluating on datasets with a known number of mark classes, we use a linear classification head to directly predict the classes of marks and group them into lines accordingly. 

\section{Experiment Results}
\subsection{Metric}
We evaluate the performance on data extraction using F1 score which is defined as 

\begin{equation}
F1= 200 \times \frac{precision \times recall}{precision + recall}
\label{eq:f1}
\end{equation}

where precision and recall are calculated under a tolerance of 5\% relative error for SYN dataset and an absolute error of 5dB for Audiogram dataset. 

When multiple lines are present, the precision and recall are calculated by iterating over all possible correspondences between detected lines and ground truth and taking the best match.

\subsection{Data Extraction on Camera Images}
We train our ChartRCNN model using a global batch size of 16 across 4 GPUs for 100 epochs. We use Adam \cite{kingma2014adam} optimizer with a learning rate of $10^{-4}$ and a weight decay of 1.0e-6. The momentum is 0.9. The learning rate is linearly scaled up in the first 3 warmup epochs, then follows a cosine decay schedule.

We perform evaluations on 4 datasets: SYN-Test, SYN-Camera, Audiogram, and Audiogram-Scanned. All models are trained using SYN dataset. For SYN-Test and SYN-Camera evaluation, we trained on the online version of the SYN dataset with infinitely many images. For Audiogram and Audiogram-Scanned, we train our model on a subset of SYN dataset. We limit the range of randomly generated data and the styles to match the audiogram. For example, the x-axis is generated in a log scale instead of a linear one as the general SYN dataset. All our methods are never trained on real camera images.

To evaluate the effectiveness of our synthetic data generation pipeline, we compared our results with models trained on labeled images in Audiogram dataset. This model uses a standard Faster-RCNN to detect and classify marks and labels. The OCR is performed through a classification head by considering a set of possible common tick labels in audiograms as separate classes. We report the results in \cref{tab:main_result}. To the best of our knowledge, there are no comparable works capable of extracting line chart data from ill-aligned non-clean images. Hence we were unable to compare against other works.
\begin{figure*}
    \centering
    \includegraphics[width=1.0\textwidth]{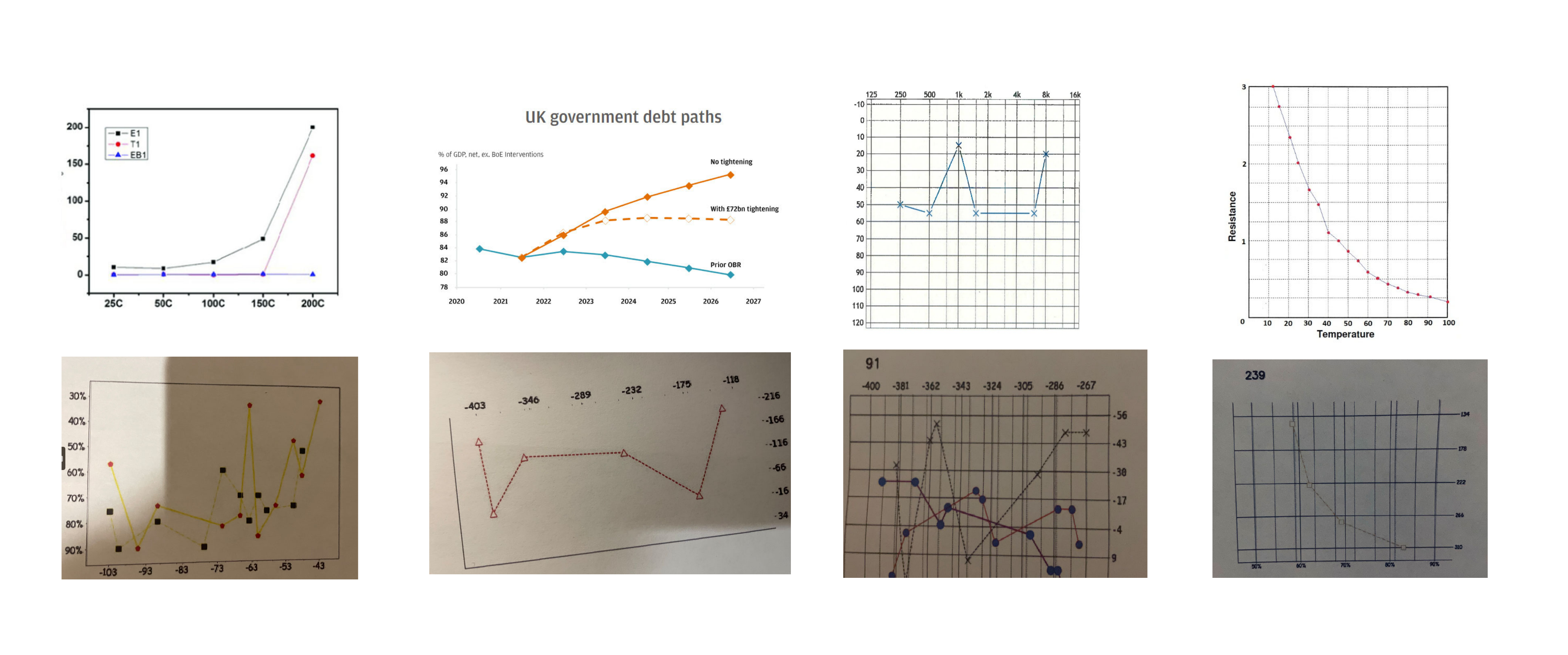}

   \caption{\textbf{SYN as a Universal Approximation.} We find several typical use-cases of line charts in different domains: 1. Material Science Paper 2. Financial Forecast 3. Audiogram 4. Specs of Electronic Components. We observe that there are some images in our randomly generated SYN-Camera dataset that share certain similarities with these images. Both images in the leftmost column contain solid lines with square and circle marks. Both images in the second most left column contain no top and right border and have a dashed line. In the third column, the style of the audiogram is an exact match for the style of a line in the bottom image. In the rightmost column, both images have lines of similar trends presented in a grid.  }
   \label{fig:test}
\end{figure*}

\begin{table}[h]
  \centering
  \begin{tabular}{ccc}
    \toprule
    Dataset & Ours & Supervised \\
    \midrule
    SYN-Test & 90.2 &  -\\
    SYN-Camera & 85.3 &  - \\
    Audiogram & \textbf{87.3} &  84.3     \\
    Audiogram-Scanned & 97.1 & \textbf{98.9}\\
    \bottomrule
  \end{tabular}
  \caption{Data Extraction on Clean and Camera Images. Numbers are F1 scores. Our model defeats the supervised baseline trained on camera photos of audiograms, and is comparable on scanned images. }
  \label{tab:main_result}
\end{table}

As is shown in \cref{tab:main_result}. Our model defeats the supervised baseline (+3.0) trained on real camera photos of audiograms without using any manually annotated data. The only human labor required is to enter a prior distribution of data ranges and line styles. Hence, it is an effective method for such tasks. Our model also achieved comparable high performance (97.1 vs 98.9) on scanned audiograms. For SYN dataset, we did not compare our method against a supervised baseline because labeling camera photos are time-consuming. We believe in using audiograms which are generated using Noah 4 \cite{himsa_noah} that are used by audiologists across the world in clinical practices can better demonstrate the effectiveness of our method in practical application scenarios. Hence we chose to label the audiograms.

\subsection{Generalizability and Limitations }

We argue that our  experiments on the SYN dataset provide good intuition on how our model generalizes on real data because of its variation in styles. To show this, we demonstrate in \cref{fig:test} that  charts from different domains are proper subsets of SYN under certain parametrization. We find line charts in different domains have corresponding images in the randomly generated SYN-Camera dataset that are reasonably similar. Hence, the SYN-Camera dataset is a good dataset to evaluate the performance of our model. This is also validated on a domain-specific dataset: Audiogram, where our method outperforms the supervised method trained on human-annotated labels. Moreover, by the law of large numbers, the model is likely to encounter charts that look more similar because these real-world charts are in the parameter space of our random generation process. To validate this, we test our model pretrained on the general SYN dataset for 200 epochs (2x schedule) without any prior knowledge of the style nor label distribution of Audiograms. The F1 score is \textbf{71.6} on Audiogram, suggesting a limited zero-shot transfer capability.

However, while the SYN dataset can approximate line charts in various domains, our method still has the following limitations. First, we use mark detection as the basis of line recognition, which is not applicable to line charts that do not have any marks. Second, we assume the chart region has distinguishable corners which can be used to perform homography estimation. As we have demonstrated, in \cref{fig:test}, a wide range of charts from different domains satisfy our assumptions. In domain-specific applications such as audiograms or spec sheets of electronic components,  charts have a uniform style and these assumptions are easily satisfied.

\section{Ablation and Analysis}
\label{sec:ablation}
In this section, we evaluate the contribution of different components to the performance of our proposed system. For all ablation experiments, we chose a shorter schedule (20 epochs) than the full evaluation (100 epochs) due to limited computing resources.  The experiments are evaluated on Audiogram Dataset.

\subsection{Transforms}
\begin{table}[h]
  \centering
  \begin{tabular}{@{}lc@{}}
    \toprule
    Transforms & F1 \\
    \midrule
    Baseline & 14.3 \\
    + Perspective & 17.7 \\
    + Color & 76.1\\
    + Blur and Noise  & 82.1\\
    + TPP & \textbf{84.8} \\
    \bottomrule
  \end{tabular}
  \caption{\textbf{Ablation Results of Different Transforms.} All proposed transforms lead to a positive increase in  F1 scores. Color seems to be the most important as it simulates different white balance settings of a camera.}
  \label{tab:ablation_transforms}
\end{table}

We employ a wide range of transforms in the data transformation process as described in \cref{sec:synthetic_data_generation} and illustrated in \cref{fig:transforms}. To evaluate the importance of each transform, we start with a baseline in which the model is trained on only clean images just like most existing works. Then we gradually add different transforms into the data generation process. The results are shown in \cref{tab:ablation_transforms}. All proposed transforms lead to a positive increase in  F1 scores among which Color has the most significant impact. We believe this is likely due to the fact that it emulates the different white balancing settings of a camera and different lighting conditions. Without such transforms, the model could easily overfit to data with a pure white background which is rather rare in real camera photos. Blur and Noise are the second most important transform because it emulates the change in focal length and real-world noises captured by camera sensors. We believe it also mitigates the overfitting problem caused by the model attending to specific local structure that is unique to certain fonts or the way in which a line or circle is drawn by the computer. By adding high-frequency noises, the local structure is perturbed.   

\subsection{Number of Images}
\begin{table}[h]
  \centering
  \begin{tabular}{@{}lc@{}}
    \toprule
    Num of Images & F1 \\
    \midrule
    100 & 76.7 \\
    1,000 & 81.4 \\
    10,000 & 81.9 \\
    \textbf{$\infty$} & \textbf{84.8}\\

    \bottomrule
  \end{tabular}
  \caption{\textbf{Ablation Results on the Number of Training Images.} The performance improves with more images. }
  \label{tab:num_of_images}
\end{table}
We employ an online dataset of infinitely many images which is rare in related works. The stochastic process also makes the training process indeterministic. To evaluate whether this is necessary, we explored alternative options where the model is only trained on a fixed number of pre-generated images. As is shown in \cref{tab:num_of_images}, the performance improves with more images. We believe that an online dataset solves the overfitting problem because it is hard to overfit a dataset with infinitely many images. Additionally, we observe no significant increase in data time. Hence our method is desirable. 

\subsection{OCR}
Unlike \cite{luo2021chartocr} which uses external OCR API or \cite{Kato2022ParsingLC} which uses a trained OCR, we fine-tune an OCR model jointly during the training. We believe this is necessary for that it can help the model learn a prior distribution of text labels present in specific types of charts, which can compensate for the additional difficulty caused by noises and perspective projections in camera photos. To validate our hypothesis, we plug in a  CRNN model pretrained for number recognition into our trained model and compare how it affects the performance. We also benchmarked a transformer-based Text Decoder proposed in \cref{sec:text_recog} against CRNN.

\begin{table}[h]
  \centering
  \begin{tabular}{@{}lcc@{}}
    \toprule
    Text Decoder & Homography Predictor & F1 \\
    \midrule
    Online CRNN & Keypoint &  \textbf{84.8} \\
    Online Transformers & Keypoint   & 81.9 \\
    Pretrained CRNN & Keypoint  & 79.4\\

    Online CRNN & Mask & 84.3 \\
    \bottomrule
  \end{tabular}
  \caption{\textbf{Ablation Results for Text Decoder and Homography Predictor.} Online CRNN has the best performance with a lead of +2.9. Both methods for homography estimation have comparable performance. Due to the complexity of the mask-based method, we picked the keypoint-based one as our final design. }
  \label{tab:ocr_ablation}
\end{table}

As shown in \cref{tab:ocr_ablation},  Online CRNN has the best performance with a lead of +2.9 against the Transformer and a lead of +5.4 against trained CNN. This confirms our hypothesis that online OCR training is necessary. While several benchmarks showed \cite{wrobel2021ocr,Li2021TrOCRTO} that Transformers tend to outperform CNN-based architecture,  they are harder to train and require more delicate tuning. This could be why we failed to observe an expected lead in Transformers. Due to limited computing resources, we were unable to explore more variants of transformer-based OCR methods.

\subsection{Homography Estimation}

We estimate homography transforms by first predicting the four corners of the chart region. We proposed a mask-based approach and a keypoint-based approach in \cref{sec:homography}. We benchmarked their performance. The results are shown in \cref{tab:ocr_ablation}. While both methods achieved similar performance, the mask-based approach requires complicated postprocessing. Hence, we find the keypoint-based approach which is more intuitive and straightforward is more desirable.

\subsection{Error Analysis}

So far all the analyses are based on the F1 score which tests the overall performance of our method as an integrated system. To further ablate the source of errors, we replace part of the model output with ground truth annotations and evaluate the performance of the remaining system.  These experiments are performed on images sampled from a manually annotated training set of the Audiogram dataset because it requires ground truth annotations of bounding boxes. This is ok because our models are never exposed to human labels. However, this set of images is not used in our other experiments because it is exposed to a supervised baseline. Hence, the numbers are not directly comparable with our other experiments. The results are shown in \cref{tab:error_analysis} 
\begin{table}[h]
  \centering
  \begin{tabular}{@{}lc@{}}
    \toprule
    Architecture & F1 \\
    \midrule
    Chart-RCNN & 86.3 \\
    + GT Detection & 87.3 \\
    + GT OCR &  87.3 \\\hline
     All GT &  100.0 \\
    \bottomrule
  \end{tabular}
  \caption{\textbf{Error Analysis}. Adding ground truth bounding boxes improves the result while further adding ground truth OCR annotations leads to no significant improvements.}
  \label{tab:error_analysis}
\end{table}

We discovered that adding ground truth bounding boxes improves the result while further adding ground truth OCR annotations leads to no significant improvements. This suggests our online fine-tuning process combined with a robust RANSAC regressor can effectively handle noises in OCR output. The results also suggest that most error comes from homography estimation. This is reasonable since we are setting a small tolerance of 5dB, and small errors in keypoint estimations may propagate to the estimated transformation matrix, which in turn translates to reprojected coordinates. This process involves inverting and multiplying several matrices and in this process, the errors may be amplified in a super-linear fashion. 

\section{Conclusion}
One of the most significant challenges of chart data extraction from camera photos is the cost to obtain annotated training samples. We proposed a customizable synthetic dataset based on a matplotlib with rich styles and can approximate chart styles in specific domains. We introduce a series of transforms that make a clean chart looks visually similar to a camera photo. We proposed Chart-RCNN, a unified architecture to extract data from camera photos of line charts. We demonstrated through experiments a cost-effective way of training such a model using no annotated data which still achieves comparable results with models trained on human-annotated data. Evaluations suggest our model is feasible for real-world applications such as audiogram interpretation. 

Ablation analysis suggests most errors come from the perspective estimation process. Our future works will attempt to incorporate some of the state-of-the-art architecture beyond 4-point-based methods. We will also explore more possible applications such as helping the visually impaired to read charts in printed documents and books.

{\small
\bibliographystyle{ieee_fullname}
\bibliography{egbib}

\begin{thebibliography}{10}\itemsep=-1pt

\bibitem{Davila2020ICPR2}
Kenny Davila, Chris Tensmeyer, Sumit Shekhar, Hrituraj Singh, Srirangaraj
  Setlur, and Venu Govindaraju.
\newblock Icpr 2020 - competition on harvesting raw tables from infographics.
\newblock In {\em ICPR Workshops}, 2020.

\bibitem{tps}
Jean Duchon.
\newblock Splines minimizing rotation-invariant semi-norms in sobolev spaces.
\newblock In Walter Schempp and Karl Zeller, editors, {\em Constructive Theory
  of Functions of Several Variables}, pages 85--100, Berlin, Heidelberg, 1977.
  Springer Berlin Heidelberg.

\bibitem{He2020MaskR}
Kaiming He, Georgia Gkioxari, Piotr Doll{\'a}r, and Ross~B. Girshick.
\newblock Mask r-cnn.
\newblock {\em IEEE Transactions on Pattern Analysis and Machine Intelligence},
  42:386--397, 2020.

\bibitem{himsa_noah}
HIMSA.
\newblock Noah system 4 – himsa.

\bibitem{huang2007extraction}
Weihua Huang, Ruizhe Liu, and C-L Tan.
\newblock Extraction of vectorized graphical information from scientific chart
  images.
\newblock In {\em Ninth International Conference on Document Analysis and
  Recognition (ICDAR 2007)}, volume~1, pages 521--525. IEEE, 2007.

\bibitem{10.1145/1284420.1284427}
Weihua Huang and Chew~Lim Tan.
\newblock A system for understanding imaged infographics and its applications.
\newblock In {\em Proceedings of the 2007 ACM Symposium on Document
  Engineering}, DocEng '07, page 9–18, New York, NY, USA, 2007. Association
  for Computing Machinery.

\bibitem{hunter2007matplotlib}
John~D Hunter.
\newblock Matplotlib: A 2d graphics environment.
\newblock {\em Computing in science \& engineering}, 9(03):90--95, 2007.

\bibitem{Jung2017ChartSenseID}
Daekyoung Jung, Wonjae Kim, Hyunjoo Song, Jeongin Hwang, Bongshin Lee,
  Bo~Hyoung Kim, and Jinwook Seo.
\newblock Chartsense: Interactive data extraction from chart images.
\newblock {\em Proceedings of the 2017 CHI Conference on Human Factors in
  Computing Systems}, 2017.

\bibitem{kafle2018dvqa}
Kushal Kafle, Brian Price, Scott Cohen, and Christopher Kanan.
\newblock Dvqa: Understanding data visualizations via question answering.
\newblock In {\em Proceedings of the IEEE conference on computer vision and
  pattern recognition}, pages 5648--5656, 2018.

\bibitem{Kahou2018FigureQAAA}
Samira~Ebrahimi Kahou, Adam Atkinson, Vincent Michalski, {\'A}kos
  K{\'a}d{\'a}r, Adam Trischler, and Yoshua Bengio.
\newblock Figureqa: An annotated figure dataset for visual reasoning.
\newblock {\em ArXiv}, abs/1710.07300, 2018.

\bibitem{Kato2022ParsingLC}
Hajime Kato, Mitsuru Nakazawa, Hsuan-Kung Yang, Mark Chen, and Bj{\"o}rn
  Stenger.
\newblock Parsing line chart images using linear programming.
\newblock {\em 2022 IEEE/CVF Winter Conference on Applications of Computer
  Vision (WACV)}, pages 2553--2562, 2022.

\bibitem{kingma2014adam}
Diederik~P Kingma and Jimmy Ba.
\newblock Adam: A method for stochastic optimization.
\newblock {\em arXiv preprint arXiv:1412.6980}, 2014.

\bibitem{kobayashi1999toward}
Ichiro Kobayashi.
\newblock Toward text based information processing: with an example of natural
  language modeling of a line chart.
\newblock In {\em IEEE SMC'99 Conference Proceedings. 1999 IEEE International
  Conference on Systems, Man, and Cybernetics (Cat. No. 99CH37028)}, volume~5,
  pages 202--207. IEEE, 1999.

\bibitem{kosemen2020multi}
Cem Kosemen and Derya Birant.
\newblock Multi-label classification of line chart images using convolutional
  neural networks.
\newblock {\em SN Applied Sciences}, 2(7):1--20, 2020.

\bibitem{Li2021TrOCRTO}
Minghao Li, Tengchao Lv, Lei Cui, Yijuan Lu, Dinei A.~F. Flor{\^e}ncio, Cha
  Zhang, Zhoujun Li, and Furu Wei.
\newblock Trocr: Transformer-based optical character recognition with
  pre-trained models.
\newblock {\em ArXiv}, abs/2109.10282, 2021.

\bibitem{luo2021chartocr}
Junyu Luo, Zekun Li, Jinpeng Wang, and Chin-Yew Lin.
\newblock Chartocr: Data extraction from charts images via a deep hybrid
  framework.
\newblock In {\em 2021 IEEE Winter Conference on Applications of Computer
  Vision (WACV)}. The Computer Vision Foundation, January 2021.

\bibitem{mou2013robust}
Wei Mou, Han Wang, Gerald Seet, and Lubing Zhou.
\newblock Robust homography estimation based on non-linear least squares
  optimization.
\newblock In {\em 2013 IEEE International Conference on Robotics and
  Biomimetics (ROBIO)}, pages 372--377. IEEE, 2013.

\bibitem{Poco2017ReverseEngineeringVR}
Jorge Poco and Jeffrey Heer.
\newblock Reverse‐engineering visualizations: Recovering visual encodings
  from chart images.
\newblock {\em Computer Graphics Forum}, 36, 2017.

\bibitem{rai2010survey}
Pradeep Rai and Shubha Singh.
\newblock A survey of clustering techniques.
\newblock {\em International Journal of Computer Applications}, 7(12):1--5,
  2010.

\bibitem{Ren2015FasterRT}
Shaoqing Ren, Kaiming He, Ross~B. Girshick, and Jian Sun.
\newblock Faster r-cnn: Towards real-time object detection with region proposal
  networks.
\newblock {\em IEEE Transactions on Pattern Analysis and Machine Intelligence},
  39:1137--1149, 2015.

\bibitem{savva2011revision}
Manolis Savva, Nicholas Kong, Arti Chhajta, Li Fei-Fei, Maneesh Agrawala, and
  Jeffrey Heer.
\newblock Revision: Automated classification, analysis and redesign of chart
  images.
\newblock In {\em Proceedings of the 24th annual ACM symposium on User
  interface software and technology}, pages 393--402, 2011.

\bibitem{Shelhamer2015FullyCN}
Evan Shelhamer, Jonathan Long, and Trevor Darrell.
\newblock Fully convolutional networks for semantic segmentation.
\newblock {\em 2015 IEEE Conference on Computer Vision and Pattern Recognition
  (CVPR)}, pages 3431--3440, 2015.

\bibitem{Shi2017AnET}
Baoguang Shi, Xiang Bai, and Cong Yao.
\newblock An end-to-end trainable neural network for image-based sequence
  recognition and its application to scene text recognition.
\newblock {\em IEEE Transactions on Pattern Analysis and Machine Intelligence},
  39:2298--2304, 2017.

\bibitem{Suzuki1985TopologicalSA}
Satoshi Suzuki and Keiichi Abe.
\newblock Topological structural analysis of digitized binary images by border
  following.
\newblock {\em Comput. Vis. Graph. Image Process.}, 30:32--46, 1985.

\bibitem{Visvalingam1990TheDA}
Mahes Visvalingam and J.~Duncan Whyatt.
\newblock The douglas‐peucker algorithm for line simplification:
  Re‐evaluation through visualization.
\newblock {\em Computer Graphics Forum}, 9, 1990.

\bibitem{wang2003cost}
Samuel~J Wang, Blackford Middleton, Lisa~A Prosser, Christiana~G Bardon,
  Cynthia~D Spurr, Patricia~J Carchidi, Anne~F Kittler, Robert~C Goldszer,
  David~G Fairchild, Andrew~J Sussman, et~al.
\newblock A cost-benefit analysis of electronic medical records in primary
  care.
\newblock {\em The American journal of medicine}, 114(5):397--403, 2003.

\bibitem{wrobel2021ocr}
Krzysztof Wr{\'o}bel.
\newblock Ocr correction with encoder-decoder transformer.
\newblock 2021.

\bibitem{zhou2001chart}
Yanping Zhou and Chew~Lim Tan.
\newblock Chart analysis and recognition in document images.
\newblock In {\em Proceedings of Sixth International Conference on Document
  Analysis and Recognition}, pages 1055--1058. IEEE, 2001.

\end{thebibliography}
}
\newpage
\setcounter{section}{0}
\renewcommand{\thesection}{A.\arabic{section}}
\setcounter{figure}{0}
\renewcommand{\thefigure}{A\arabic{figure}}
\setcounter{table}{0}
\renewcommand{\thetable}{A\arabic{table}}
\setcounter{equation}{0}
\renewcommand{\theequation}{A\arabic{equation}}
\setcounter{algorithm}{0}
\renewcommand{\thealgorithm}{A\arabic{algorithm}}
\section*{Appendix}
\section{Implementation Details}
\subsection{Keypoint-based Homography Estimation}
\begin{figure}[h]
    \centering
    \includegraphics[width=0.8\linewidth]{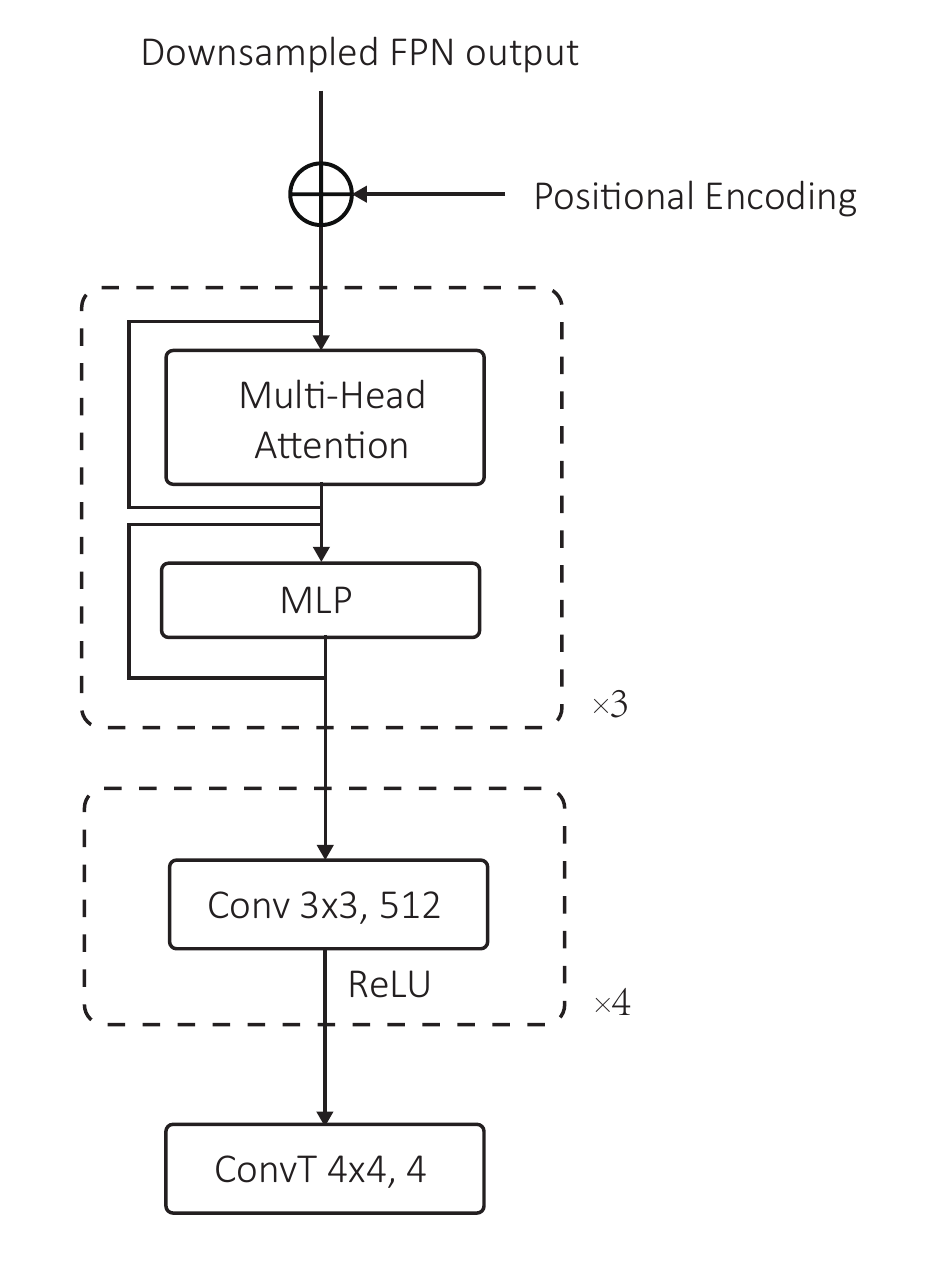}
    \caption{\textbf{Architecture of Keypoint Head }. The output of ResNet-FPN is downsampled to a 14x14 feature map. 2D Positional Encoding is then added. We consider the feature map as a sequence of length 196 and pass it through three transformer layers whose Multi-Head Attention Block has 8 heads and a dimension of 256. The output of transformer layers is then reshaped back to a feature map. It is then passed through 4 convolution layers with kernel size 3 and ReLU activation. The final output is generated by a ConvTranspose layer of  kernel size 4. }
    \label{fig:keypoint_head}
\end{figure}

We use a modified version of keypoint prediction head of \cite{He2020MaskR}. Since the local structures of the four corners of a chart are similar, and some charts do not have a border at corners, we add an attention mechanism so that the model can access global positional information but mere local structures.  The output of the ResNet-FPN backbone is downsampled to a 14x14 feature map. This is considered as a sequence of length 196. We add 2D positional encoding and pass the sequence through 3 transformer layers. The sequence is then reshaped back to a 14x14 feature map, which is then passed through 4 additional convolution layers. To get the final output of a 4-channel keypoint heatmap, we use a ConvTranspose layer of kernel size 4. The exact architecture is shown in \cref{fig:keypoint_head} 
\subsection{Mask-based Homography Estimation}
\begin{figure}[t]
    \centering
    \includegraphics[width=0.8\linewidth]{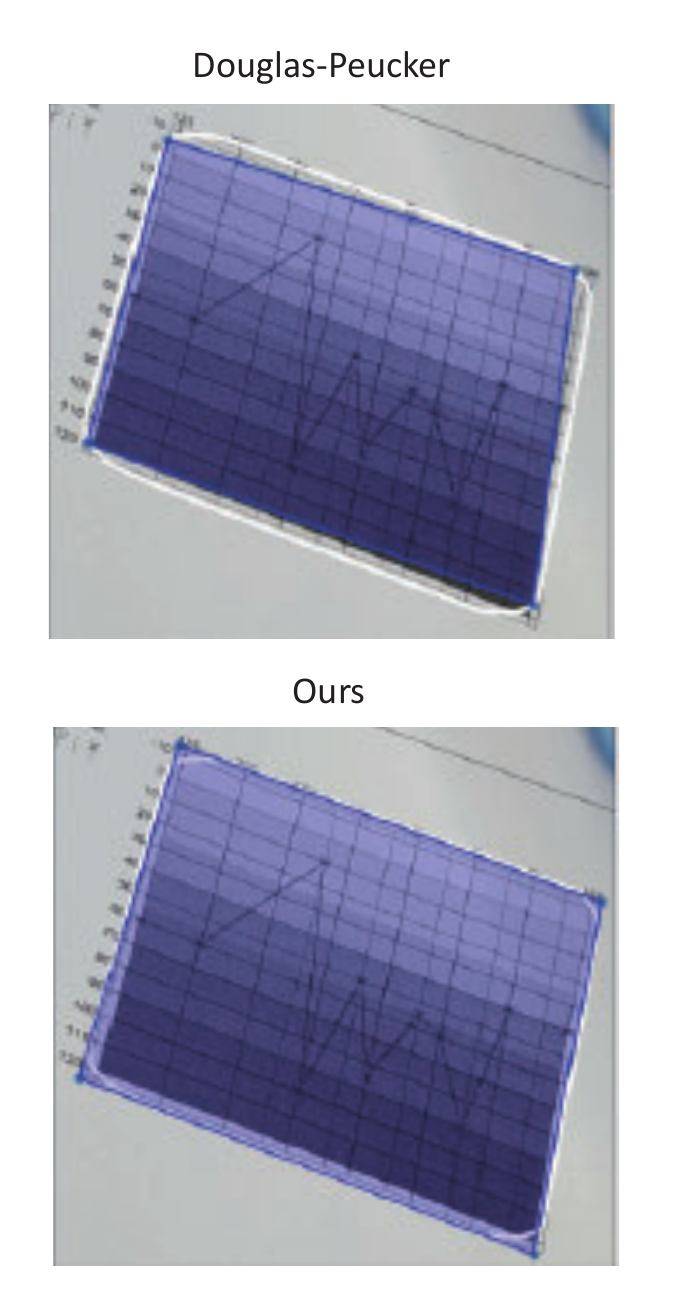}
    \caption{\textbf{Comparison Between Ours and Douglas-Peucker }. The Douglas-Peucker is typically used to simplify polygons. However, it only reduces a polygon to a subset of the original vertices. In this example, the white line is the polygon generated by predicted masks, and the colored regions are reduced quadrilaterals. The four corners of the chart are actually outside the original polygon, hence Douglas-Peucker fails to recover them. }
    \label{fig:dp_mask}
\end{figure}
We use the same mask prediction head following \cite{He2020MaskR}. However, it is non-trivial to recover a quadrilateral from predicted binary masks because of noises in predictions. Typically, such tasks are performed by Douglas‐Peucker Algorithm \cite{Visvalingam1990TheDA} which reduces complicated polygons to simple ones. However, it is flawed because the vertices of the simplified polygon are always a subset of the vertices of the original polygon. This can give sub-optimal approximations when the four corners of a chart are not the vertices of the predicated mask. As shown in \cref{fig:dp_mask}, our method overcomes this problem.

The details of this procedure are described in \cref{alg:cap}. Given an input binary mask $M$, we first convert it to a set of contours $C$. Then we find $c\in C$ which is the contour with the largest enclosed area. Let $h$ be its convex hull, we set the threshold of Douglas‐Peucker as $0.5\%$ of the circumference of $h$ and perform standard  Douglas‐Peucker. This gives us a reduced polygon $c_{approx}$. Let $E$ be the set of its edges. We want to find a subset $E_4$ containing 4 edges that will be the edges of the final reduced quadrilateral. We initialize $E_4$  as an empty set, then repetitively adding the longest edge in $E$ that is not too close to an existing edge in $E_4$. In particular, we consider two edges $e_1,e_2$ as too close if the angle between them is less than $20$ degrees and the minimum distance between three points (two ends and the midpoint) of one line and an arbitrary point on the other line is less than a $1/28$ of image dimension. With four edges, we find all their intersections and filter out those outside the image region. We then sort the remaining points by the angle of the vector from the centroid of all these points to a specific point. This recovers an ordered set of four points that can be used to solve a projective transformation matrix. 

\begin{algorithm}
\caption{Homography Estimation on Binary Mask}\label{alg:cap}
\begin{algorithmic}
\Require $M \in \{0,1 \}^{H \times W}  \,\, \, \,   \text{Binary Mask}$
\Ensure $p_1,p_2,p_3,p_4 \in \mathbb{R}^2  \,\, \, \,   \text{Keypoints}$
\State $C \gets \textsc{cv2.FindContours}(M) $
\State $c \gets \textsc{LargestContourByArea}(C) $
\State $h \gets \textsc{cv2.ConvexHull}(c) $
\State $\epsilon \gets 0.005 \cdot length(h)$
\State $ c_{approx} \gets \textsc{DouglasPeucker}(c,\epsilon)$
\State $ E \gets \textsc{GetEdges}(c_{approx})$
\State $ E_4 \gets \{\}$
\While {$|E_4| < 4$} 
\State $ e \gets  \textsc{GetLongestLine}(E) $
\State $ E_4.\text{add}(e) $
\State $ E.\text{pop}(e)  $
\State $ E.\text{removeCloseLines}(e)  $
\EndWhile
\State $ P\gets \textsc{GetIntersections}(E_4)$ 
\State $ p_1,p_2,p_3,p_4 \gets \textsc{SortByAngle}(P)$ 
\end{algorithmic}
\end{algorithm}

\subsection{Mark Grouping Head}
\begin{figure}[h]
    \centering
    \includegraphics[width=0.8\linewidth]{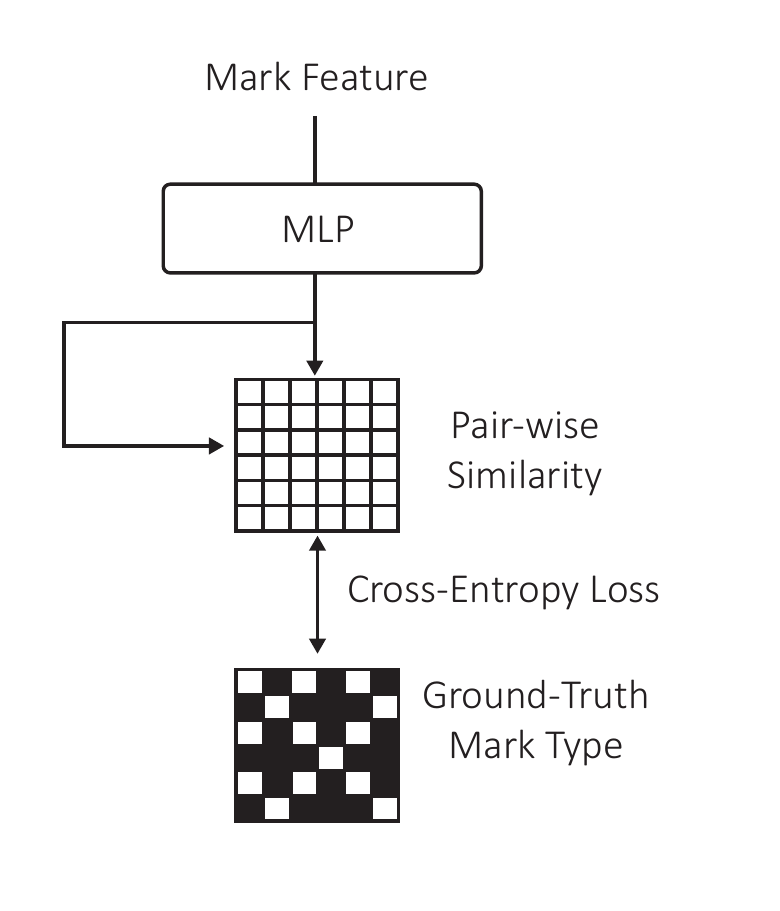}
    \caption{\textbf{Architecture of Mark Grouping Head }. The mark features are first projected to a different space by an MLP. We then compute the pair-wise cosine similarity of all mark features. We consider the resulting matrix as log-likelihood of an random variable indicating if two marks belong to the same line and apply cross-entropy loss on it following \cref{eq:loss}. }
    \label{fig:mark_grouping}
\end{figure}

We illustrate the architecture of our Mark Grouping Head in \cref{fig:mark_grouping}. The mark features are first projected to a different space by an MLP. We then compute the pair-wise cosine similarity of all mark features. We consider the resulting matrix as the log-likelihood of a random variable indicating if two marks belong to the same line and apply cross-entropy loss on it following \cref{eq:loss}. In \cref{eq:loss}, we introduce a temperature hyperparameter $\tau$ because the cosine similarity is bounded in the interval $[-1,1]$. $\tau$ scales this interval to a larger range. In batched training, we apply a mask and the loss is only applied to the similarity between marks in the same image. We also masked out the diagonal entries of the similarity matrix because they are trivial. In order to add more flexibility to the output range so that the similarity is not always centered at zero, we add a learnable bias term that is added to the cosine similarity matrix. 

\section{F1 Score under Different Error Tolerance}
\begin{figure*}[t]
    \centering
    \includegraphics[width=0.8\linewidth]{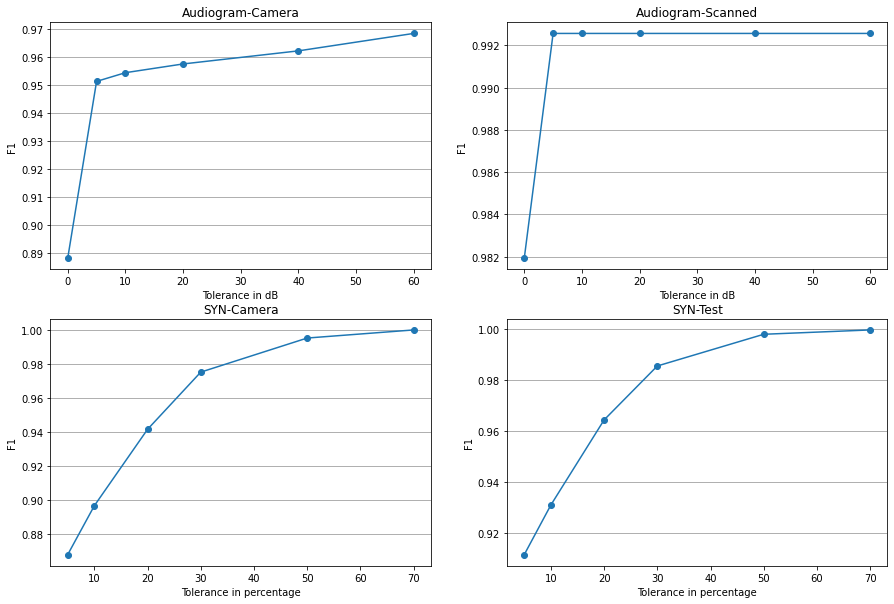}
    \caption{\textbf{F1 under difference tolerance}. To further evaluate our method, we plot the F1 score of our model on different datasets across different error tolerance. In audiogram datasets, the tolerance is measured by dB. In SYN dataset, the tolerance is measured by percentage. }
    \label{fig:acc}
\end{figure*}
To further evaluate our method, we plot the F1 score of our model on different datasets across different error tolerance in \cref{fig:acc}. We noticed that in Audiogram-Camera, we can achieve an F1 score of $95.1$ with only a $5$dB tolerance, which is half the length of the gap between two consecutive ticks. This suggests most errors are within a margin of $5$dB. In SYN datasets, the F1 continues to increase as the error tolerance gets larger, suggesting a wider distribution of errors. This is likely caused by the varying chart styles in the dataset with different levels of difficulty. 
\section{More Visualization of Datasets}
\begin{figure*}[t]
    \centering
    \includegraphics[width=0.8\linewidth]{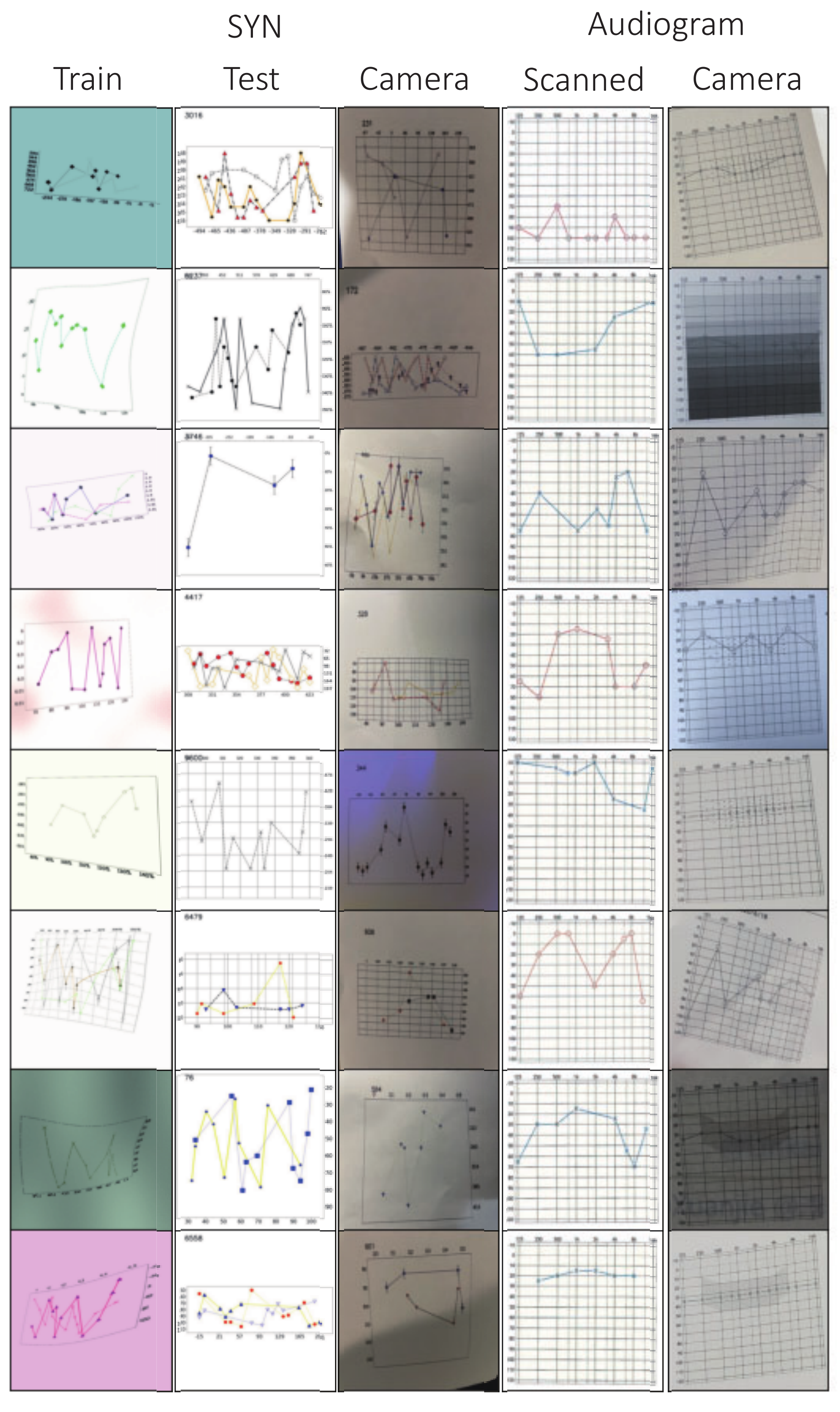}
    \caption{\textbf{More Visualization of Samples in Datasets.} The leftmost three columns are SYN-Train, SYN-Test, and SYN-Camera. The rightmost two columns are Audiogram-Camera and Audiogram Scanned. }
    \label{fig:more_data}
\end{figure*}

In \cref{fig:more_data} we provide more visualization of samples in the datasets we used. In particular, columns SYN-Train, SYN-Camera, and Audiogram-Camera demonstrate the similarity of synthetic data and camera photos.  

\end{document}